\documentclass[10pt]{article}
\usepackage[letterpaper]{geometry}
\usepackage{hicss}
\usepackage{times}
\usepackage[none]{hyphenat}
\usepackage{url}
\usepackage{latexsym}
\usepackage{indentfirst}
\usepackage{graphicx}
\graphicspath{{images/}}
\usepackage{caption}
\usepackage{subcaption}
\usepackage[
    style=apa,
  ]{biblatex}

\usepackage{booktabs}
\usepackage{multirow}
\usepackage{xcolor}
\usepackage{cuted}  
\usepackage{float}
\usepackage{afterpage}
\usepackage{amsmath}

\makeatletter
\renewcommand{\verbatim@font}{\normalfont\ttfamily\bfseries\color{black}}
\makeatother
\newsavebox{\vbatbox}

\graphicspath{{images/}}

\title{Restrict, Don't Retrain: Inference-Time VLM Guidance for Zero-Shot Aerial Segmentation}

\date{}

\begin{document}
\maketitle

\begin{abstract}

Global welfare often depends on the correct interpretation of aerial and satellite imagery. Acting on such imagery (mapping flooded ground, crop extent,
  or damaged infrastructure) demands pixel-level segmentation to ensure perfect class localization. Pretrained general
  foundation models, when applied directly, often miss important features and cannot always find all the classes belonging to a given scene, overlooking smaller objects that
  matter most. We use a single consumer-grade GPU running a vision-language model (VLM) to supply this missing guidance, improving segmentation while
  producing structured, auditable evidence that drives the result and can be inspected on its own. We fuse three approaches: the frozen foundation model
  that labels every pixel, and two queries to a VLM, one to choose the classes that matter, and one to locate the small objects the base model misses. 
  Evaluating across four aerial datasets, we see consistent gains at each stage where the base model is competent.


\end{abstract}

\subsubsection*{Keywords:}

aerial imagery, vision-language models, zero-shot segmentation, open vocabulary segmentation, auditable inference

\section{Introduction}
\label{sec:intro}

Many practical decisions now rest on interpreting aerial and satellite imagery.
Disaster-response teams triage flooded or damaged areas, agricultural planners
map land cover, and infrastructure crews inspect roads, roofs, and power
corridors. In each case a raw image must be turned into structured information
that can be acted on. This is carried out via semantic segmentation: assigning every pixel an object class.

The usual path to good aerial segmentation is to build a domain-specific
foundation model, training a large network from scratch on very large amounts of
in-domain data (\cite{wen2023VLM}). This is costly, as it needs labeled data,
compute infrastructure, machine-learning expertise, and a separate
model for each domain. When the set of classes that matters changes over time,
the model goes stale and the cost returns.

We argue this is unnecessary. A general-purpose foundation model already
encodes the relevant classes. What it lacks, at inference time, is knowledge of
which part of its large vocabulary applies to the image in front of it, and where
the small but important objects are. The problem is one of \emph{selection} and
\emph{guidance}, not of model architecture or training-data scale.

Lighter-weight alternatives such as open-vocabulary, vocabulary-free, and
inference-time methods have also been explored (\cite{ghiasietal2022}), but in specialized domains they
struggle to reliably select the right classes for each image.

This paper presents a workflow that supplies that guidance by bringing together
two models that cooperatively counteract the failure mode of the other. The first is the frozen foundation model
that labels every pixel but treats all classes as equally likely. The second is a secondary
vision-language model (VLM) that looks at each image and returns a short, structured
reading of the scene: which classes are present, and where the small objects sit.
The VLM never paints pixels itself. It works as an interpretive layer, turning a
raw image into readable cues that steer and patch the base model. We query it
twice, through two inference-time mechanisms:
\begin{itemize}
    \item \textbf{Automated Class Weighting (ACW)}: the VLM picks the handful of
          classes that actually appear in the image, and those classes' scores
          are boosted before the model makes its per-pixel decision
          (Section~\ref{sec:acw}).
    \item \textbf{Minority Class Identification (MCI)}: the VLM is asked a second
          time to draw boxes around the small objects the weighted classifier
          still misses, and those boxes are filled in and merged back
          (Section~\ref{sec:mci}).
\end{itemize}

Neither model is reliable on its own. The base model has broad pixel coverage but
weak class discipline; the VLM has strong scene judgment but cannot produce a pixel
map. Putting them together lets each cover the other's blind spot, so the pair
reaches a more reliable result than either one gives alone.

We treat the workflow as a deliberate design and study it across four aerial
datasets, in the design-science tradition. From the results we draw three design
principles that carry to other specialized domains: (i) restricting a general
model's vocabulary is preferable to training a separate domain-specific model;
(ii) weighting the majority classes a scene is built from improves coherence;
(iii) the small classes a dense model drops can be recovered by asking the VLM
where they are.

\section{Related Work}
\label{sec:related}

Remote sensing (RS) has shifted toward transformer-based vision-language methods.
A survey of VLMs for RS (\cite{wen2023VLM}) shows a clear trend toward large,
RS-specific models such as RS5M and GeoRSCLIP (\cite{zhang2023RS5M}) and RemoteCLIP
(\cite{liu2024remoteclip}), each trained as a separate model from scratch at large
cost in data, energy, and hardware.

These models are meant for ``in-the-wild'' imagery: aerial data the model was not
trained on, where the task is to label objects and landscapes the model has never
seen. This zero-shot segmentation (ZSS) needs a class label for every pixel,
including classes absent from training (\cite{bucher2019zeroshot}).
Transformer-based methods (\cite{ding2022decoupling}) rely on
large in-domain pretraining and lose accuracy in new domains unless they are
fine-tuned, or even retrained. Benchmarks such as MESS
(\cite{blumenstiel2023whatamess}) measure this transfer gap and have become a
reference point for zero-shot evaluation.

Open-vocabulary segmentation (OVS) (\cite{ghiasietal2022}) lets a user name any
class at inference time, and recent work extends CLIP (\cite{radford2021learning})
with mask-adapted heads to bring open-vocabulary capabilities into segmentation
(\cite{liang2023openvocabulary}). Vocabulary-free methods
(\cite{reichard2025vocabulary}) let the model invent its own class names, but in
specialized domains this produces inconsistent names and unpredictable output.

One reason teams build domain-specific models is the belief that the model's
weights must change, which is almost as costly as building the model from scratch.
A lighter option is to adapt a model's outputs at inference time instead of its
weights; for example, \textcite{cheng2021perpixelclassificationneedsemantic}
address class imbalance by adjusting per-pixel logits.

Most work on class imbalance in segmentation acts at training time, through
resampling or loss weighting; in aerial imagery, large changes in object scale
make the imbalance worse (\cite{elhagry2022investigatingchallengesclassimbalance}).

Using language models to enumerate visual features has emerged alongside
instruction-following models. \textcite{kao2025think} show that
chain-of-thought reasoning supports fine-grained scene description, and
\textcite{liu_visual_ins_tuning2023_neurips} show that multimodal LLMs can act
as image-aware annotators.


Beyond segmentation itself, there is a broader shift in decision
support, where large language and vision-language models are used not as standalone
predictors but as components inside larger workflows, turning unstructured inputs
into structured features that downstream tools can act on (\cite{rahimiOR}).
Such workflows work by fusing information gleaned from each element, supporting the long-standing practice of combining
several heterogeneous sources to reach a more reliable result than any one of them
gives alone (\cite{fusionSurvey}). Transparency questions can also be more readily addressed if
analysts can identify the reasonong that led to a particular decision.

By treating the workflow
as a designed artifact in the design-science
tradition (\cite{hevner2004design,sein2011action}), reusable design
principles can be leveraged over reporting from a "black-box" system (\cite{ribeiro2016lime}).
The transparency is built in via structured outputs a reviewer, who also knows the inputs, can inspect. The faithfulness concerns raised for chain-of-thought explanations (Barez et al., 2025) therefore do not apply here: we never treat the reasoning narrative as the explanation; the structured outputs it produces are what we inspect and act on.
\section{Design and Methods}
\label{sec:design}

\begin{figure*}[!ht]
    \centering
    \includegraphics[width=1\linewidth]{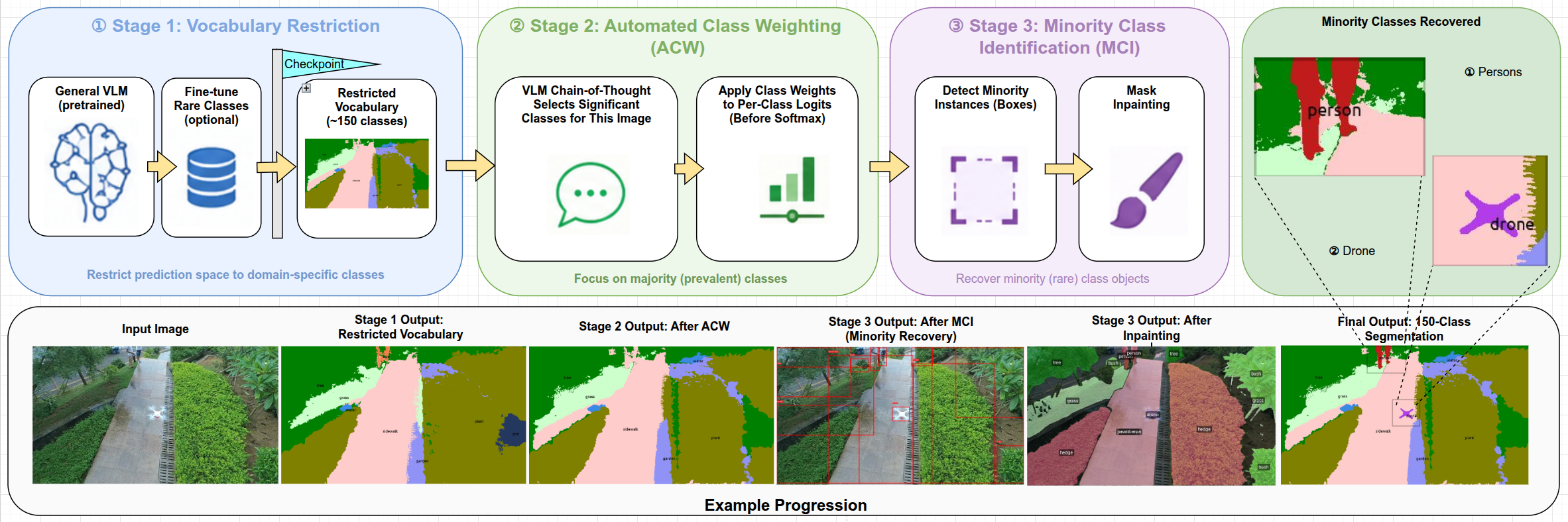}
    \caption{The three-stage workflow. Top row: an aerial image is processed through Stage~1 (vocabulary restriction to a curated set of about 150 classes), Stage~2 (automated class weighting, where a VLM chain-of-thought selects the significant classes for the image and boosts their per-class logits), and Stage~3 (minority class identification, where the VLM locates small objects as bounding boxes that are then mask-inpainted), producing the final 150-class segmentation. Bottom row: a worked example showing the output after each stage, with minority classes such as persons and a drone recovered in Stage~3.}
    \label{fig1:pipeline}
\end{figure*}

Our method splits the work between two frozen models: a general foundation model
that produces dense per-pixel labeling over a curated 150-class vocabulary, and a
vision-language model, queried twice, that supplies the per-image guidance the dense
model lacks: which classes to weight (Stage~2) and where small, rare objects sit
(Stage~3). Nothing is retrained, and every VLM decision is an inspectable list or
box. The three subsections detail each stage in turn. Figure~\ref{fig1:pipeline}
shows the full workflow alongside a worked example of the per-stage progression.

\subsection{Stage 1: Baseline Segmentation}
\label{sec:resda}
General-purpose foundation models hold a great deal of semantic knowledge from
large-scale pretraining, including many domain-specific concepts
(\cite{radford2021learning,wen2023VLM}). We tap this in two ways. First, we
restrict the inference vocabulary: we start from the 150-class ADE20K vocabulary
(\cite{zhou2017ade20k}) that the base model already predicts and hand-adjust it toward
aerial scenes, replacing general classes with ones that characterize infrastructure,
natural features, vehicles, land use, and specialized objects. Second, we fine-tune
on each dataset's annotated classes of interest, ranging from six (UDD6) to
twenty-four (DroneSeg), as anchors. The remaining classes are found zero-shot
through relationships already present in the base model's embeddings. Aerial
categories are common enough to be well represented in pretraining, so this
fine-tuning is rarely decisive here; we include it because the same blueprint
transfers to domains such as medicine, chemistry, and materials science, where
genuinely novel classes appear regularly and fine-tuning anchors become essential.

\subsection{Stage 2: Automated Class Weighting (ACW)}
\label{sec:acw}
In plain zero-shot inference every class competes for every pixel. Classes that
are not in the image add noise and crowd out the ones that are. A person glancing
at an aerial image can name the five to ten classes that appear; holding the
model to those classes sharply improves the result.

Let $\mathcal{C}$ denote the 150-class domain vocabulary. For each input image
$I$, we identify a \emph{significant class set} $\mathcal{B}_I \subset \mathcal{C}$.
We define per-class weights:
\begin{equation}
w_c = \begin{cases} W_{\text{boost}} & \text{if } c \in \mathcal{B}_I \\ W_{\text{neutral}} & \text{otherwise} \end{cases}
\label{eq:weights}
\end{equation}
where $W_{\text{boost}} = 100$ and $W_{\text{neutral}} = 0$ (or $1$ to retain a
soft fallback). These weights are applied multiplicatively to per-pixel logits
before softmax:
\begin{equation}
  \begin{split}
  \ell^{(\text{w})}_{i,j,c} &= w_c \cdot \ell_{i,j,c}, \\
  P_{i,j,c} &= \frac{\exp(\ell^{(\text{w})}_{i,j,c})}{\sum_{k} \exp(\ell^{(\text{w})}_{i,j,k})}, \\
  \hat{c}_{i,j} &= \arg\max_c P_{i,j,c}
  \end{split}
  \label{eq:acw}
  \end{equation}
No model weights are changed; ACW works with any VLM that exposes per-class
logits. It also has a useful side effect: focusing the output on a small,
image-specific class set reduces the ambiguity that would otherwise force us to
tile large images. The output of this stage is a complete per-pixel segmentation:
the base model labels every pixel from the weighted logits, while the VLM supplies
only the class set and weights, with no spatial information of its own. Localization
here is entirely the base model's; bounding boxes do not enter until Stage~3.

\begin{figure*}[t!h]
\centering
\begin{lrbox}{\vbatbox}\begin{minipage}{0.97\textwidth}
{\footnotesize
\begin{verbatim}
You will serve as an agent for a language-based image segmentation model. Your task is
to describe a given image using chain of thought, with as many details as possible to
help the model understand it. Target objects may have multiple layers, be blocked by
other objects, or be embedded in their surroundings. Pay special attention to water,
sky, clouds, walls, etc. The first questions cover the overall style or background;
then work down to fine detail (e.g., items a person is holding). Distinguish trees from
other vegetation. Include anything visible, even if it is not prominent.

Output format:
  - <Question 1>: <Answer 1>.
  - <Question 2>: <Answer 2>.
  ... each pair implies one step of the chain of thought.
After the first pass, review the image again in a second pass to discover missed items
or reassess, then add a second output.

Classes must be selected from the 150-class class_list. Return a list of every noun
from that list that you find in the image, with a careful percentage for each. If
something is unidentifiable or not on the list, call it "other" and give its percentage
too. Do not skip signs, posts, or other markers.
\end{verbatim}
}
\end{minipage}\end{lrbox}\fbox{\usebox{\vbatbox}}\\
\vspace{1ex}
\begin{lrbox}{\vbatbox}\begin{minipage}{0.97\textwidth}
{\footnotesize
\begin{verbatim}
- <Question 1>: What is the overall layout and style of the image?
  <Answer 1>: An aerial top-down view of an urban campus or institutional complex. It
  features multiple outdoor basketball courts, surrounding buildings with modern
  designs, paved parking areas, sidewalks, and small patches of vegetation...

- <Question 4>: Are there any vehicles visible?
  <Answer 4>: Yes. Three white vans are parked along the edge of the basketball courts,
  and another is on the right side of the image. Numerous bicycles are parked in
  designated racks, arranged neatly in rows.

- <Question 7>: Are there any signs, posts, or markers?
  <Answer 7>: Yes. Multiple vertical posts, likely light poles or signage supports,
  appear along the periphery of the courts and near the buildings. One sign is visible
  near the central building entrance, though its content is illegible...

[... the chain of thought continues over a second pass, then the response ends with the
required structured output: a boosted_classes list and an override_weights dictionary
with every weight set to 100.]
\end{verbatim}
}
\end{minipage}\end{lrbox}\fbox{\usebox{\vbatbox}}
\captionof{figure}{(Top): The VLM prompt for automated class selection (Stage~2), 
lightly edited for length and readability. The Stage~3 bounding-box prompt shares this structure. 
(Bottom): A representative VLM response (snippet, edited for length). Full responses can
run to several thousand tokens. The chain-of-thought catches small objects such as the
vans, bicycles, posts, and an illegible sign, the minority classes Stage~3 later
recovers. The response closes with the structured list and weight dictionary the
downstream steps consume. Thanks to \textcite{kao2025thinksegmenthighqualityreasoning} for inspiring the first part of the prompt.}
\label{fig:prompt_response}
\end{figure*}

ACW needs the class set $\mathcal{B}_I$ before inference. Hand-labeling it is
accurate but defeats the purpose of zero-shot deployment, so we obtain it from
the VLM. A two-part prompt (Figure~\ref{fig:prompt_response}) asks the model to (i) reason step by step about the
scene at both global and local levels, (ii) choose classes only from the
150-class vocabulary, and (iii) return a Python-ready list and weight dictionary
so the output parses cleanly. Here the VLM is doing the interpretive work: it
turns the image into a readable list of classes and weights that a person can
check. The result feeds straight into ACW (Equation~\ref{eq:acw}, above):
\begin{equation}
\mathcal{B}_I = \text{LM\_prompt}(I,\, \mathcal{C}).
\label{eq:classsel}
\end{equation}
We judge class-selection quality with precision, recall, and F1 score against the
ground-truth class set $G(I)$ present in image $I$:
\begin{equation}
  \begin{split}
  \text{Prec}  :=& \ \ \text{Precision}(I) = \frac{|\mathcal{B}_I \cap G(I)|}{|\mathcal{B}_I|}, \\
  \text{Rec}  := &\ \  \text{Recall}(I) = \frac{|\mathcal{B}_I \cap G(I)|}{|G(I)|}, \\
  \text{F1}(I) = &\ \ 2 \cdot \frac{\text{Prec} \cdot \text{Rec}}{\text{Prec} + \text{Rec}}.
  \end{split}
  \label{eq:f1}
\end{equation}

\subsection{Stage 3: Minority Class Identification (MCI)}
\label{sec:mci}
Even after Stage 2, some classes the VLM reports as present never appear in the
segmentation: the base model assigns them no pixels. We call these
\emph{minority classes}: signs, vehicles, windows, and similar objects that occupy
few pixels but matter to the scene. They are small and sparse enough that the
model's per-pixel signal for them is too weak to win at any location, and class
weighting can only scale that signal, not manufacture it where the model produced
almost none. Stage 2 already labels most pixels correctly; MCI is a focused fix for
the classes Stage 2 cannot reach.

MCI is the workflow's second query to the same VLM. Stage 2 asked what is in the
scene; Stage 3 asks where the small objects are. We prompt the VLM to return
bounding boxes for the minority-class objects it sees, in image coordinates. Each box is
then filled in to a precise pixel mask by a class-agnostic segmenter
(\cite{kirillov2023segment}) and merged back into the Stage 1+2 result. When the merge meets a minority class absent from the Stage 1 baseline,
it paints the class in directly; when it meets a class already present, it updates
those pixels only if the new mask scores better by per-class IoU. Keeping the only
judgment calls inside the VLM's readable outputs, and leaving the fill step
mechanical, is what makes the result easy to trace back to its source.

Figure~\ref{fig:prompt_response}  makes the two VLM queries concrete:
the first shows an example prompt, the second a snippet of a representative response,
including the chain-of-thought and the structured class list and weights it emits.

\section{Experimental Setup}
\subsection{Datasets}
\label{sec:datasets}
We evaluate our approach on four aerial benchmark datasets:
\textbf{UAVid} (\cite{lyu2020uavid}) (8 classes, urban drone video),
\textbf{Aeroscapes} (\cite{nigam2018ensemble}) (11 classes, diverse aerial views),
\textbf{DroneSeg} (\cite{ji2024pptformer}) (24 classes, diverse geography), and
\textbf{UDD6} (\cite{chen2018udd}) (6 classes, campus and urban rooftops).

\subsection{Evaluation Protocol}
\label{sec:eval-protocol}
A core methodological choice is rigorous cross-dataset evaluation for zero-shot
transfer. For each dataset pair (A, B): fine-tune on A, test zero-shot on B. We
map the restricted vocabulary of 150 domain-specific classes to each dataset's
ground-truth class set for mIoU calculation. We note that ground truth is used only
for computing evaluation metrics (such as mIoU), and not for weighting, prompt
construction, or minority class identification (MCI). The primary evidence for our central
claim is the \emph{consistency} of $\Delta$mIoU across stage transitions and
across datasets, not the absolute mIoU on any single dataset. Absolute mIoU is a
conservative lower bound: ground-truth annotations frequently use catch-all
classes (\textit{Clutter}, \textit{Unlabeled}, \textit{Other}) for pixels the
annotators did not label, so correct predictions in those regions are penalized
as false positives. We remove all such catch-all classes from all predictions. This
bound is tightest where the ground truth is coarsest: our restricted 150-class
vocabulary encodes distinctions the benchmark labels cannot express, so the metric
penalizes the model for predicting more finely than the annotation can score. UDD6 is
the extreme case, with only six coarse classes, so the richest part of the mapping has
nowhere to land and correct fine predictions are counted as errors.

\subsection{Hardware}
\label{sec:experimental-setup}
All experiments are conducted on an NVIDIA RTX A6000 GPU with memory capped to
16\,GB to simulate consumer-hardware constraints (the original target was a
16\,GB consumer laptop GPU). The fine-tuning hyperparameters we used are: Adam, lr $= 5 \!\times\!
10^{-5}$, weight decay $0.1$, batch size 16, and 2k iterations. Stage 1 fine-tuning
completes in 1 to 2 hours per dataset on a single GPU. To account for VLM
response stochasticity in the Stage 2 class selection, we run three independent
trials (seeds $s\!=\!0,1,2$) for each combination of dataset and configuration, and report mean
$\pm$ standard deviation; statistical significance versus the unweighted baseline
is assessed via paired $t$-test. Runtime is dominated by VLM API latency in Stages 2 and 3, which is highly variable
and capped at 20 minutes per call. The VLM API accounts for roughly 90\% of the
pipeline's runtime, leaving local computation (base-model inference and instance
segmentation) about 10\%.

\subsection{VLM Screening for Bounding Box Generation and Inpainting}
\label{sec:vlm_comparison}
Several VLMs were considered for the MCI bounding-box step in early experiments,
including GPT-4o, Claude Sonnet, and Gemini. None produced bounding boxes whose
locations corresponded usefully to objects in aerial imagery: predicted boxes were
displaced from their targets by magnitudes large enough to render the subsequent
fill step ineffective. Qwen-VL was the only model that produced spatially-grounded
predictions suitable for the MCI step. We did not pursue a quantitative comparison
because the gap was qualitatively decisive at the screening stage; candidates that 
cannot place a box within the same region as the target object cannot be ranked
meaningfully against one that can. Spatial grounding accuracy in commercial VLMs
on aerial imagery remains an open problem. Once the class-labeled bounding boxes have been
defined, they are inpainted.

\section{Results}
\label{sec:results}

\subsection{Qualitative Results}
\label{sec:qualitative}

\begin{figure}[h!]
\centering
  \includegraphics[width=1\columnwidth]{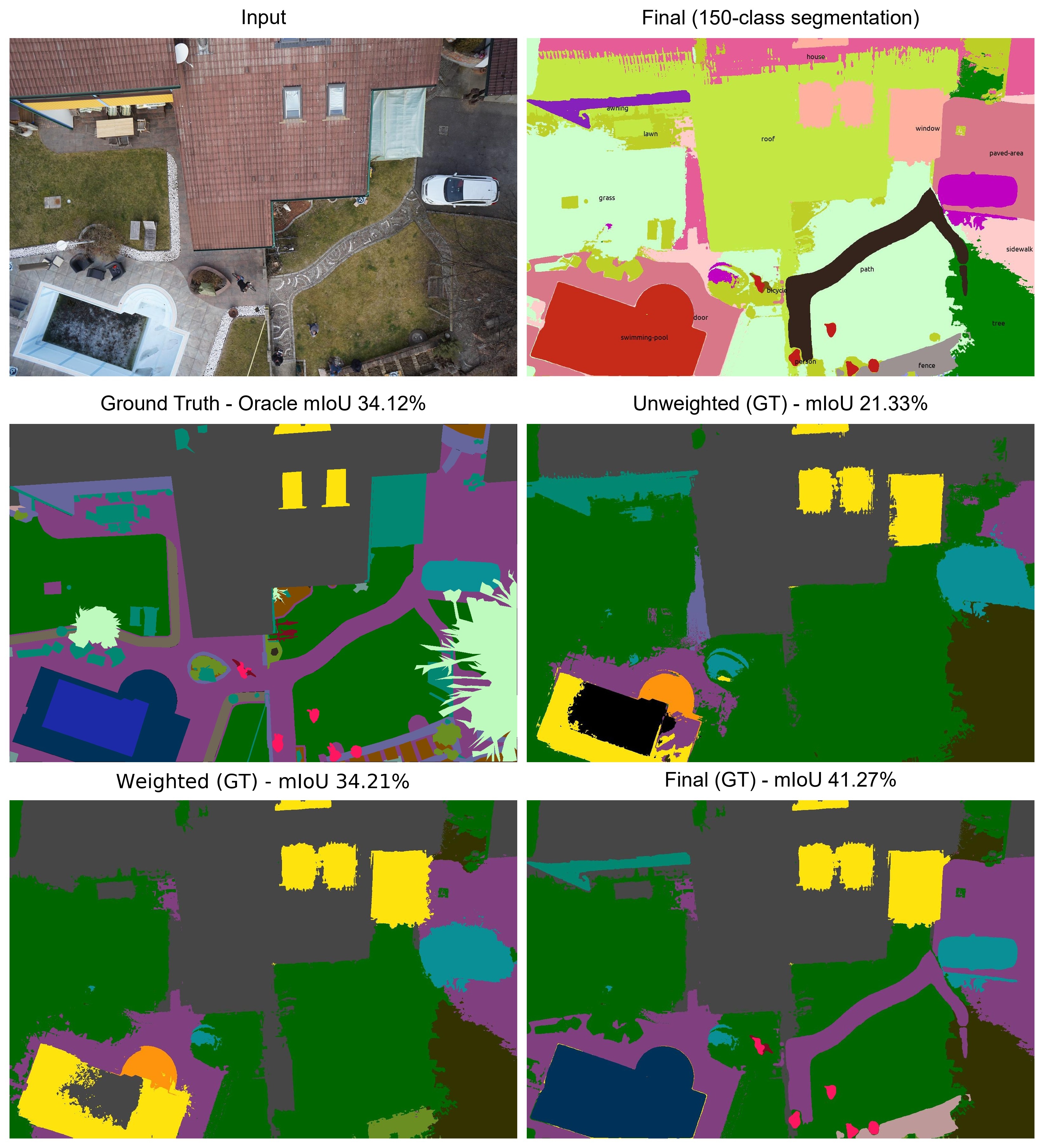}
  \captionof{figure}{DroneSeg, image~514. Top row: input image and final 150-class segmentation. Bottom two rows: ground-truth annotation (Oracle mIoU shown) and the pipeline outputs at each stage, unweighted (Stage~1), weighted (Stage~2), and final merged (Stage~3), all in ground-truth class colors.}
  \label{fig:droneSeg}
\end{figure}

\begin{figure}[t]
    \centering
    \includegraphics[width=1\columnwidth]{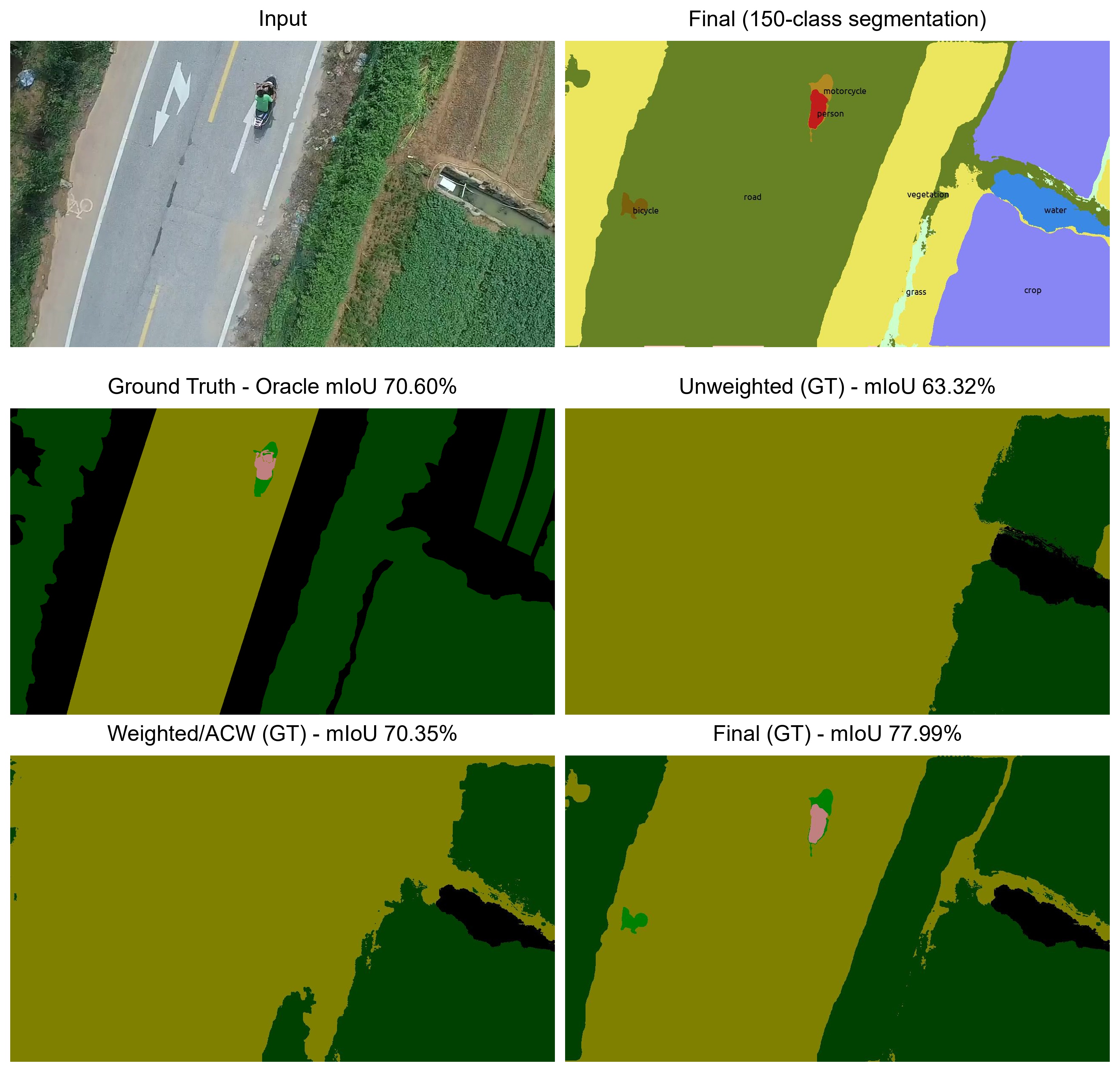}
    \caption{Aeroscapes, image~001101\_001. Top row: input image and final 150-class segmentation. Bottom two rows: ground-truth annotation (Oracle mIoU shown) and the pipeline outputs at each stage, unweighted (Stage~1), weighted (Stage~2), and final merged (Stage~3), all in ground-truth class colors.}
    \label{fig:aero}
\end{figure}

\begin{figure}
\centering
  \includegraphics[width=\columnwidth]{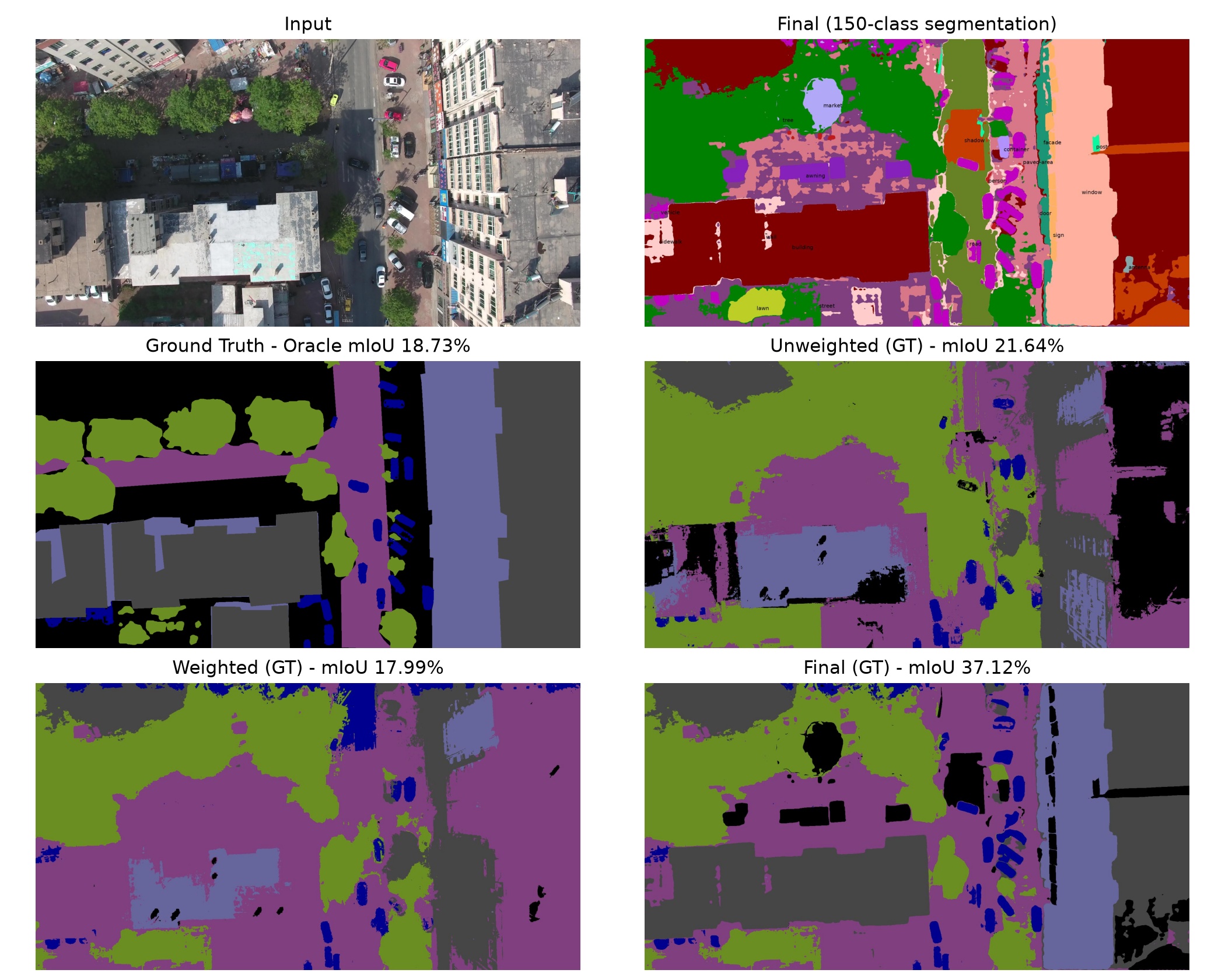}
  \captionof{figure}{UDD6, image~001361. Top row: input image and final 150-class segmentation. Bottom two rows: ground-truth annotation (Oracle mIoU shown) and the pipeline outputs at each stage, unweighted (Stage~1), weighted (Stage~2), and final merged (Stage~3), presented in ground-truth class colors.}
  \label{fig:udd6}
\end{figure}

\begin{figure}[t]
    \centering
    \includegraphics[width=1\columnwidth]{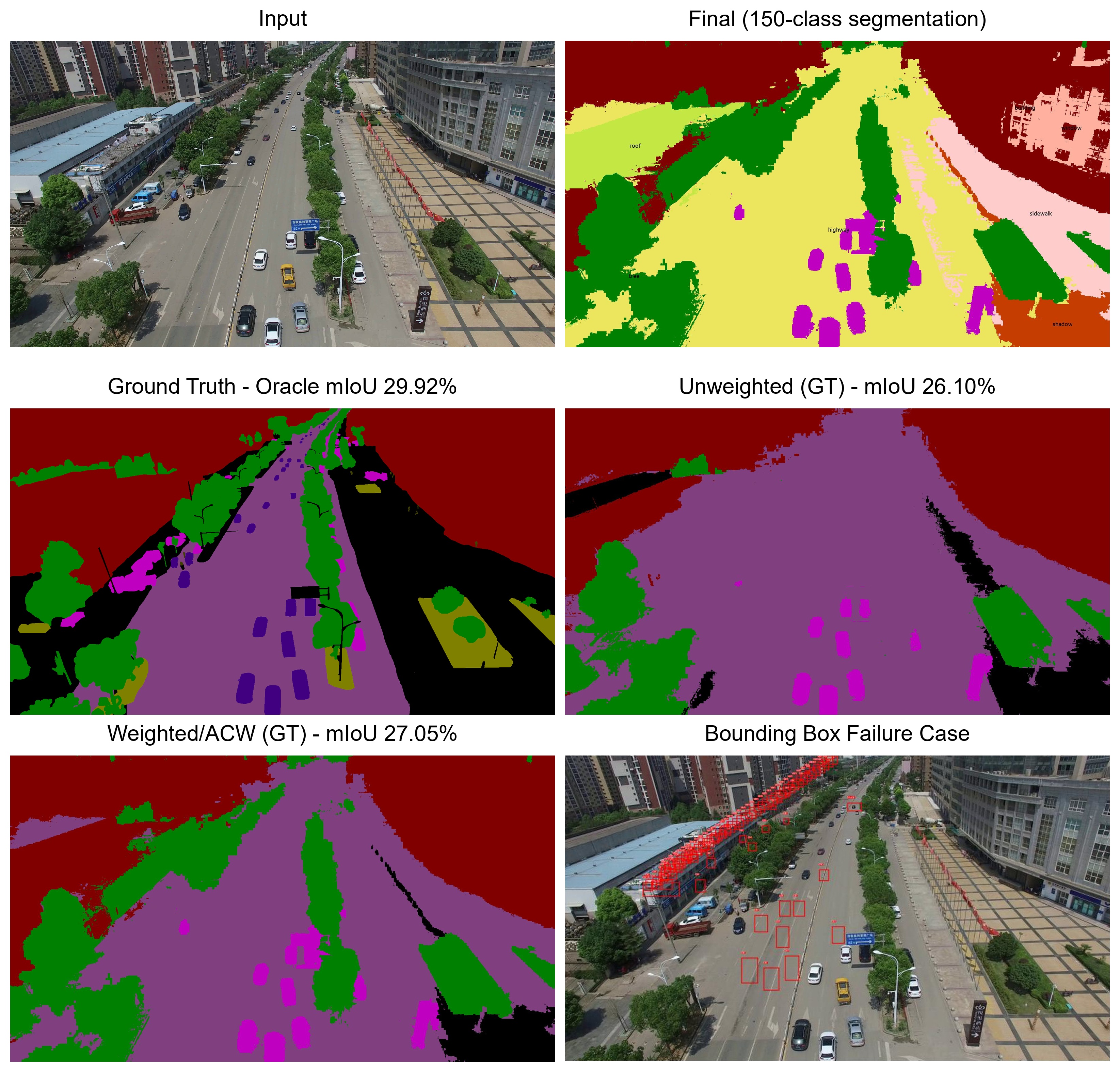}
    \caption{UAVid, image~uavid0000000012. Top row: input image and final 150-class segmentation. Bottom two rows: ground-truth annotation (Oracle mIoU shown) and the pipeline outputs at each stage, unweighted (Stage~1), weighted (Stage~2), and in ground-truth class colors. The final image in the bottom row illustrates how bounding box calculations do not currently compensate for the oblique, angled viewpoint of UAVid imagery: the predicted boxes drift off their targets and into the distance, so the inpainting step has nothing correctly placed to fill, and Stage 3 cannot improve on the Stage 2 result.}
    \label{fig:uavid}
\end{figure}

Our design targets the operator or researcher in the field, at the edge, without
access to backend GPU power.  The data of interest, often returned via
satellite or drone, is completely new ``in-the-wild'' imagery that must be as
accurate as possible. Figures~\ref{fig:droneSeg},~\ref{fig:aero},~\ref{fig:udd6}, and~\ref{fig:uavid} illustrate
the segmentation progression on each dataset. 

The effects of
automated class weighting (ACW) versus minority class identification (MCI) depend
on several variables, including landscape type, the mix of minority and majority
classes, and the field and angle of view. Our design captures these variations
successfully in most images. A representative suburban scene is shown in
Figure~\ref{fig:droneSeg}, Figure~\ref{fig:aero} shows a rural scene, while Figures~\ref{fig:udd6} and~\ref{fig:uavid} show a contrasting urban
scene where minority classes (cars) dominate.

UDD6 segments well because its images are almost nadir (straight-down) drone
views (see Figure~\ref{fig:udd6}), while UAVid sometimes fails because its images
vary in field of view and angle (see Figure~\ref{fig:uavid}), conditions where VLMs systematically struggle to
ground objects (\cite{egocentric_bias_vlm_2026}). 

All examples show the merged final stage exceeding the per-image Oracle. The
Oracle is the best mIoU achievable with perfect class-vocabulary
selection. The final merged stage exceeds it because Stage 3, MCI, recovers
\emph{minority classes}: classes the VLM sees in the image but the segmentation
algorithm cannot pixel-localize. In Stage 3, for each such class, a prompted VLM
returns a bounding box that is filled in, and this output is merged with Stage~2
for the final segmented output.

Data for the DroneSeg (Figure~\ref{fig:droneSeg}) and Aeroscapes (Figure~\ref{fig:aero}) datasets has been constructed for the purpose
of model training: the range and distance of objects does not vary to the extremes that complicate UDD6's dense urban clutter or UAVid's oblique perspectives.
Objects appear in moderate, consistent scales and ranges so both the class weighting and the bounding box step behave well, and each stage of the 
pipeline contributes a clean gain.

Figure~\ref{fig:udd6} makes this concrete. On this UDD6 image, ACW does not help:
the weighted result (17.99\% mIoU) falls below the unweighted baseline (21.64\%).
The full pipeline still reaches 37.12\%, because MCI recovers the small classes that
dominate this scene and delivers almost all of the gain. ACW and MCI thus address
complementary failure modes: when one does not help, the other can still carry the
result.

Figure~\ref{fig:uavid} offers a particularly clear view into the challenges of relying on even state-of-the-art approaches to augment segmentation: ill
  equipped to handle the oblique, steeply angled viewpoints that aerial video like UAVid presents, the VLM returns bounding boxes that drift away from the
  vehicles and people they should mark, floating toward the horizon. Because those boxes are misplaced, the inpainting step has nothing correctly
  localized to recover, and Stage 3 collapses onto Stage 2; this is why UAVid's Stage 3 is left undefined in our tables (Section~\ref{sec:significance}). 

\subsection{Quantitative: Progressive mIoU across All Three Stages}
The primary evidence that our lightweight, local workflow produces reliable
results within a constrained environment is the typical consistency of
$\Delta$mIoU across stage transitions. Tables~\ref{tab:prog1}, ~\ref{tab:prog2},
~\ref{tab:prog3}, and ~\ref{tab:prog4} report mIoU at each stage for four separate
underlying checkpoints (Section~\ref{sec:eval-protocol}). The meaningful signal is that the gains are large and
significant on the datasets where the backbone is competent (Aeroscapes and DroneSeg);
UDD6 and UAVid stay near flat, for the distinct reasons taken up in
Section~\ref{sec:handoff}. All values are cumulative.

We take a checkpoint after fine-tuning the VLM using one of the four datasets (Figure~\ref{fig1:pipeline} Stage 1). A checkpoint comprises the current (trained) state of the frozen foundation  model, including its weights. If finetuning was carried out the weights are updated. We have finetuned on four different datasets, therefore we have four different checkpoints. Each of these checkpoints include the theoretical unique classes added to the curated set of classes that identify the domain, a total of 150.

\subsection{Statistical Significance}
\label{sec:significance}
To check whether the per-stage gains could be due to chance, we ran a paired
$t$-test on the per-image mIoU change for each stage transition on each dataset
(Aero-checkpoint output, same per-image methodology as Table~\ref{tab:prog1}).
Table~\ref{tab:significance} reports the results. $N$ is the number of test images
with paired data; mean $d$ is the mean per-image mIoU change in percentage points.

\begin{table}[H]
\centering
\captionsetup{justification=raggedright,singlelinecheck=false}
\caption{Paired $t$-test on per-image mIoU change relative to the unweighted baseline (Stage~1).}
\label{tab:significance}
\footnotesize
\setlength{\tabcolsep}{4pt}
\begin{tabular}{llrrr}
\toprule
Dataset & Transition & $N$ & mean $d$ & $p$ \\
\midrule
Aero.  & Stage 1 $\to$ 2 & 35 & $+7.02$ & $<0.0001$ \\
Aero.  & Stage 2 $\to$ 3 & 35 & $+7.65$ & $<0.0001$ \\
UAVid  & Stage 1 $\to$ 2 & 35 & $+0.94$ & n.s. \\
UAVid  & Stage 2 $\to$ 3 & -- & --      & --   \\
DSeg   & Stage 1 $\to$ 2 & 35 & $+6.63$ & $<0.0001$ \\
DSeg   & Stage 2 $\to$ 3 & 35 & $+6.20$ & $<0.0001$ \\
UDD6   & Stage 1 $\to$ 2 & 35 & $-0.24$ & n.s. \\
UDD6   & Stage 2 $\to$ 3 & 31 & $+3.74$ & $<0.01$ \\
\bottomrule
\end{tabular}
\end{table}

The difference between the unweighted baseline and class weighting (ACW), that is,
between Stage 1 and Stage 2, is highly significant on Aeroscapes and DroneSeg
($p<0.0001$) and not significant on UAVid or UDD6. The difference between Stage 1 and
Stage 3 (the full pipeline) is likewise highly significant on Aeroscapes ($+15.07$) and
DroneSeg ($+14.68$), $p<0.0001$, and not significant on UAVid ($+1.18$) or UDD6
($-1.89$). The UAVid rows are computed over the $N{=}19$ images on which the VLM was
successfully able to locate boxes; multiple viewing angles and perspectives prevent
reliable localization on the remainder (Sections~\ref{sec:qualitative}
and~\ref{sec:vlm_comparison}).

\subsection{Complementary Recovery Across Landscapes}
\label{sec:handoff}

\begin{table*}[t]
\centering
\begin{minipage}[t]{0.48\textwidth}
\centering
\caption{Aero checkpoint\\{\normalfont\footnotesize Stage~1 fine-tuned on Aeroscapes, evaluated across all four datasets.}}
\label{tab:prog1}
\footnotesize
\setlength{\tabcolsep}{3pt}
\begin{tabular}{lcccc}
\toprule
Stage & UAVid & Aero. & DSeg & UDD6 \\
\midrule
Stage 1: Unweighted & 26.02 & 62.99 & 52.54 & 27.52 \\
Stage 2: Weighted $w\!=\!100$ & 26.46 & 70.36 & 59.45 & 26.90 \\
Stage 3: Minority Classes &27.20 & 78.06 & 67.22 & 25.63 \\
\bottomrule
\end{tabular}
\end{minipage}\hfill
\begin{minipage}[t]{0.48\textwidth}
\centering
\caption{DroneSeg checkpoint\\{\normalfont\footnotesize Stage~1 fine-tuned on DroneSeg, evaluated across all four datasets.}}
\label{tab:prog2}
\footnotesize
\setlength{\tabcolsep}{3pt}
\begin{tabular}{lcccc}
\toprule
Stage & UAVid & Aero. & DSeg & UDD6 \\
\midrule
Stage 1: Unweighted & 28.45 & 59.40 & 50.51 & 23.64 \\
Stage 2: Weighted $w\!=\!100$ & 28.70 & 70.41 & 59.03 & 23.37 \\
Stage 3: Minority Classes & 29.88 & 76.67 & 66.15 & 25.07 \\
\bottomrule
\end{tabular}
\end{minipage}

\vspace{1.5em}

\begin{minipage}[t]{0.48\textwidth}
\centering
\caption{UAVid checkpoint\\{\normalfont\footnotesize Stage~1 fine-tuned on UAVid, evaluated across all four datasets.}}
\label{tab:prog3}
\footnotesize
\setlength{\tabcolsep}{3pt}
\begin{tabular}{lcccc}
\toprule
Stage & UAVid & Aero. & DSeg & UDD6 \\
\midrule
Stage 1: Unweighted & 24.79 & 62.06 & 51.26 & 25.45\\
Stage 2: Weighted $w\!=\!100$ & 24.92 & 69.68 & 58.55 & 25.45 \\
Stage 3: Minority Classes & 25.90 & 77.40 & 67.23 & 24.89 \\
\bottomrule
\end{tabular}
\end{minipage}\hfill
\begin{minipage}[t]{0.48\textwidth}
\centering
\caption{UDD6 checkpoint\\{\normalfont\footnotesize Stage~1 fine-tuned on UDD6, evaluated across all four datasets.}}
\label{tab:prog4}
\footnotesize
\setlength{\tabcolsep}{3pt}
\begin{tabular}{lcccc}
\toprule
Stage & UAVid & Aero. & DSeg & UDD6 \\
\midrule
Stage 1: Unweighted & 14.68 & 53.35 & 49.52 & 17.02 \\
Stage 2: Weighted $w\!=\!100$ & 16.78 & 68.36 & 58.20 & 20.13 \\
Stage 3: Minority Classes & 17.90 & 76.96 & 66.10 & 22.08 \\
\bottomrule
\end{tabular}
\end{minipage}
\end{table*}

The two prompts are designed so that neither must succeed everywhere; which one
carries an image is set by the landscape. ACW sharpens the dominant land-cover
classes a scene is built from; MCI recovers the small, sparse objects a per-pixel
segmenter drops. Where majority structure is strong, ACW leads and MCI refines;
where small objects dominate, MCI carries it. The workflow never depends on a single
mechanism being correct, the practical payoff of pairing two complementary models.

This runs against a common assumption in decision support, that a language model
should wrap a conventional solver rather than replace it (\cite{rahimiOR}). MCI
shows the opposite can hold in a bounded regime: for the small, sparse classes a
per-pixel segmenter systematically drops, the VLM does not assist the solver; it
stands in for it, supplying the labels the solver could not produce.

Tables~\ref{tab:prog1} through \ref{tab:prog4} show this happening. On Aeroscapes and DroneSeg, ACW is the
larger early gain and is highly significant (Stage 1$\to$2, $p<0.0001$,
Table~\ref{tab:significance}), and MCI adds a comparable gain on top. On UDD6 the
pattern is reversed. Class weighting produces no gain: the Stage 1 to Stage 2
difference is $-0.62$ and not significant, and the full Stage 1 to Stage 3 difference is
$-1.89$, also not significant. This is a granularity artifact of the annotation set, not
a method failure: UDD6's six coarse classes leave the richest part of the 150-class
vocabulary nowhere to land, so minority recovery that is real on individual dense scenes
(Figure~\ref{fig:udd6}) is lost once the score is averaged over the image. A more finely
annotated ground truth would recover it.

The handoff matters most where mIoU is least trustworthy. Minority classes are
small in pixel count, so the metric barely registers them even when recovered
perfectly, and the ground-truth catch-all labels (Section~\ref{sec:eval-protocol})
penalize correct minority predictions as false positives. The dense urban regime,
where MCI takes over, is thus where mIoU undercounts most, which is why the
qualitative panels, not the mIoU column, are the honest evidence for that recovery.
UAVid stays flat for a different reason: its multiple viewing angles and perspectives
keep the VLM from locating boxes on most images, so MCI has little to add, and where it
does locate them the recovered cars and people are too few pixels to move a mean
dominated by large background classes.

\section{Ablation Studies}
\label{sec:ablations}

To isolate the contribution of individual design choices, and to characterize how
sensitive the method is to the selected VLM hyperparameters, we perform a set of
ablations covering the weight value used in ACW and the class-selection strategy
that drives it.

\subsection{ACW Weight Value}
\label{sec:ablation_weight}
Table~\ref{tab:ablation_weight} reports Stage~1 mIoU as the boost weight
$W_\text{boost}$ is swept from $w\!=\!0$ (the unweighted baseline) to $w\!=\!500$.
Prediction quality is stable across the sweep, with no large gains beyond moderate
weighting; we adopt $w\!=\!100$ as the production setting for all other
experiments. The shape of the sweep is informative: nearly all the gain appears at
the first nonzero weight, and mIoU is then flat out to $w\!=\!500$. The exception is
UDD6, where ACW is inert: the unweighted baseline ($w\!=\!0$, 27.52) exceeds every
weighted value (26.7--26.9), so restriction neither helps nor hurts there. The benefit
therefore comes from restricting prediction to the selected classes, not from the
magnitude of the boost; once the boost is large enough for those classes to win the
per-pixel argmax, larger values change little. The method is thus robust to this
hyperparameter, and $w\!=\!100$ sits safely in the flat region.


\begin{table}[thb]
\centering
\caption{BCI weight ablation (Stage 1 mIoU, \%).}
\label{tab:ablation_weight}
\footnotesize
\setlength{\tabcolsep}{3pt}
\begin{tabular}{lcccccc}
\toprule
Dataset & $w\!=\!0$ & $w\!=\!10$ & $w\!=\!50$ & $w\!=\!100$ & $w\!=\!200$ & $w\!=\!500$ \\
\midrule
UAVid & 30.73 & 32.77 & 32.48 & 32.05 & 32.40 & 32.48 \\
Aero. & 63.41 & 71.32 & 71.24 & 70.66 & 71.16 & 71.14 \\
DSeg & 51.29 & 59.45 & 59.40 & 59.08 & 59.38 & -- \\
UDD6 & 31.55 & 34.17 & 34.46 & 35.76 & 34.39 & 34.40 \\
\bottomrule
\end{tabular}
\end{table}

\subsection{Class Selection Strategy}
\label{sec:ablation_selection}
Table~\ref{tab:ablation_selection} addresses whether the improvement depends on which classes ACW boosts, or simply on the act of boosting. Holding the boost weight fixed at $W_\text{boost}=100$, we vary only how the boosted class set is chosen and compare three strategies: Random, which boosts $k=30$ classes drawn at random for each image (a lower bound, no real selection); VLM (Qwen), our automated method, in which the model picks the classes from the image; and Oracle, which boosts exactly the classes the ground truth says are present (an upper bound no automated method can beat). For Random and VLM we report mean $\pm$ standard
deviation over three seeds ($s\!=\!0,1,2$); Random's variance comes from which $k$
classes are sampled from the 150-class vocabulary, while VLM's reflects Qwen's
sampling-based decoding. Oracle and Unweighted are deterministic single-value
entries. The VLM selection matches the oracle and clears the random baseline,
which indicates that the automated class discovery, not merely the act of
weighting, drives the improvement. The gap over random is dataset-dependent: it
widens on the weaker-backbone and smaller-vocabulary datasets, where selection
precision matters most, and it shrinks where the backbone already predicts the
common aerial classes well without boosting.

Two further patterns stand out. Random selection is not merely weaker but, on UAVid
and UDD6, worse than no weighting at all, and highly variable across seeds: boosting
the wrong classes actively suppresses the right ones, so weighting helps only when
the selected set is correct. Selection quality, not weighting itself, is what
matters. The VLM matches the oracle to within about 0.6 mIoU on every dataset and
exceeds it on UAVid (27.14 vs.\ 25.40), despite the oracle's access to ground-truth
class sets, which is consistent with ground-truth
labels being a conservative target (Section~\ref{sec:eval-protocol}): the VLM
sometimes names useful classes the ground-truth set omits. Its low variance shows the
automated selection is stable, unlike random.

\begin{table}[thb]
\centering
\caption{Class-selection strategy comparison (Stage 1 mIoU, \%).}
\label{tab:ablation_selection}
\resizebox{1\columnwidth}{!}{

\footnotesize
\setlength{\tabcolsep}{3pt}
\begin{tabular}{lcccc}
\toprule
Strategy & UAVid & Aero. & DSeg & UDD6 \\
\midrule
Unweighted & 30.73 & 63.41 & 51.29 & 31.55 \\
Random & 19.77$\pm$7.55 & 69.89$\pm$1.62 & 57.97$\pm$2.23 & 23.76$\pm$5.33 \\
Oracle & 29.92 & 70.60 & 59.22 & 31.40 \\
VLM (Qwen) & 31.74$\pm$0.30 & 70.95$\pm$0.53 & 59.03$\pm$0.17 & 36.04$\pm$0.39 \\
\bottomrule

\end{tabular}
}
\end{table}


\section{Conclusion}
\label{sec:conclusion}

We have shown that pairing a frozen foundation model with a vision-language model,
used only to read and describe the scene, produces more scene-coherent segmentation,
measurably and consistently, without any domain-specific training. The improvement
is not specific to the aerial domain or to particular benchmark choices:
$\Delta$mIoU is positive and highly significant across the datasets where the backbone
is competent; where it is flat, the limit is the benchmark or the viewpoint, not the
method (Section~\ref{sec:handoff}). The base model, with a small amount of targeted
tuning, already knew the relevant classes; the VLM simply forced it to apply that
knowledge to the scene in front of it.

The approach can be added to any local, general foundation model without domain-specific
pretraining and without access to more than a single consumer-grade GPU. If there
are rare, unique, or new classes that pretraining may have missed, simple
fine-tuning on those classes is enough.

The two mechanisms introduced here, automated class weighting and minority class identification, rest on the same idea: a VLM that
reads the image can supply structured cues to guide a frozen foundation model,
without ever doing the pixel work itself. ACW keeps the output tuned to the
classes a prompt selects automatically, and MCI repairs the residual failure mode
of minority-class under-representation by pairing VLM-derived (class, bounding-box)
pairs with a class-agnostic fill step. Together they provide clearly better
zero-shot segmentation, both quantitatively (mIoU) and qualitatively. Each
mechanism is independently applicable and backbone-agnostic, and because every
judgement is stated in the VLM's readable outputs, each step of the result can be
traced and checked. This traceability is what makes the workflow suitable as a
component in a larger decision-making system, where the reasoning behind an
output matters as much as the output itself.

Our future work centers on increasing the accuracy of localization, a known, and challenging problem in VLM transformer architecture. A vision-language model can readily name the classes and objects in a scene, but cannot yet localize them reliably; this gap, not class discovery, is the principal obstacle to using a VLM as a primary source of object and material identification. Extending the workflow to other specialized domains, such as medical or industrial imaging, remains a longer-term aim.

\printbibliography

\end{document}